\documentclass[10pt,twocolumn,letterpaper]{article}

\usepackage{cvpr}
\usepackage{times}
\usepackage{url}
\usepackage{newtxtext,newtxmath}
\usepackage{graphicx}
\usepackage{amsmath}
\usepackage{amssymb}
\usepackage{subfigure}
\usepackage{hyperref}
\usepackage{wrapfig}

\usepackage{bm}
\usepackage{color}
\usepackage{comment}
\newcommand{\cHiro}[1]{{\color{black}{#1}}}

\cvprfinalcopy 

\def\cvprPaperID{****} 

\begin{document}

\title{Would Mega-scale Datasets Further Enhance Spatiotemporal 3D CNNs?}

\author{Hirokatsu Kataoka, Tenga Wakamiya, Kensho Hara, Yutaka Satoh\\
National Institute of Advanced Industrial Science and Technology (AIST)\\
Tsukuba, Ibaraki, Japan\\
{\tt\small \{hirokatsu.kataoka, tenga.wakamiya, kensho.hara, yu.satou\}@aist.go.jp}
}

\maketitle

\begin{abstract}
   How can we collect and use a video dataset to further improve spatiotemporal 3D Convolutional Neural Networks (3D CNNs)? In order to positively answer this open question in video recognition, we have conducted an exploration study using a couple of large-scale video datasets and 3D CNNs.
  In the early era of deep neural networks, 2D CNNs have been better than 3D CNNs in the context of video recognition. Recent studies revealed that 3D CNNs can outperform 2D CNNs trained on a large-scale video dataset. However, we heavily rely on architecture exploration instead of dataset consideration. Therefore, in the present paper, we conduct exploration study in order to improve spatiotemporal 3D CNNs as follows: 
  (i) Recently proposed large-scale video datasets help improve spatiotemporal 3D CNNs in terms of video classification accuracy. We reveal that a carefully annotated dataset (e.g., Kinetics-700) effectively pre-trains a video representation for a video classification task. 
  (ii) We confirm the relationships between \#category/\#instance and video classification accuracy. The results show that \#category should initially be fixed, and then \#instance is increased on a video dataset in case of dataset construction.
  (iii) In order to practically extend a video dataset, we simply concatenate publicly available datasets, such as Kinetics-700 and Moments in Time (MiT) datasets. Compared with Kinetics-700 pre-training, we further enhance spatiotemporal 3D CNNs with the merged dataset, e.g., +0.9, +3.4, and +1.1 on UCF-101, HMDB-51, and ActivityNet datasets, respectively, in terms of fine-tuning.
  (iv) In terms of recognition architecture, the Kinetics-700 and merged dataset pre-trained models increase the recognition performance to 200 layers with the Residual Network (ResNet), while the Kinetics-400 pre-trained model cannot successfully optimize the 200-layer architecture. The codes and pre-trained models used in the paper are publicly available on the GitHub\footnote{\url{https://github.com/kenshohara/3D-ResNets-PyTorch}}.
\end{abstract}

\section{Introduction}

Video recognition, which includes human action recognition and motion representation, is an active field and is based on the greatly developed image recognition with convolutional neural networks (CNNs). Video recognition is said to be more difficult than still image recognition because a video consists of an image sequence that changes slightly in every frame, in addition to the difficulties of still image recognition. The field of video recognition is being developed in terms of network architecture and larger-scale video dataset construction.

\begin{figure}[t]
\begin{center}
   \includegraphics[width=1.0\linewidth]{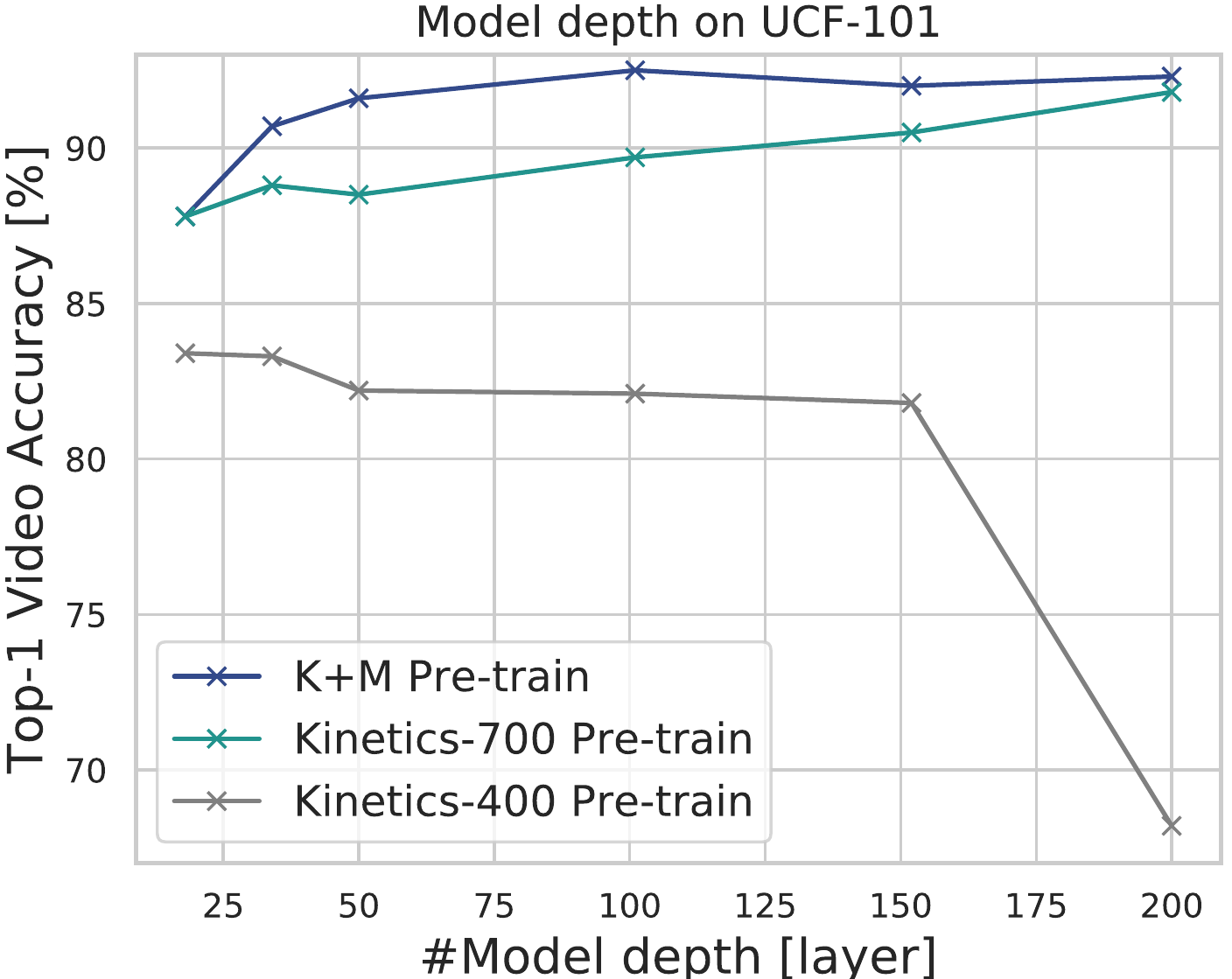}
   \caption{Though the Kinetics-400 pre-trained model is saturated/decreased the fine-tuning rates along with the model layer increase, the Kinetics-700 and merged Kinetics-700 and Moments in Time (MiT) pre-trained models have further improved the accuracy on UCF-101 validation.  According to the results, we can confirm that the increase in the dataset scale allows us to enhance spatiotemporal 3D CNNs.}
\label{fig:depth_ucf}
\end{center}
\vspace{-20pt}
\end{figure}

In recent video recognition research, we mainly have two options, i.e., 2D and 3D CNNs, which are methods for processing a video volume with a convolutional kernel. Starting from CNN+LSTM (e.g., \cite{DonahueCVPR2015}) as a baseline model, 2D CNNs have been used in Two-stream ConvNets~\cite{SimonyanNIPS2014}, which assign RGB and optical flow sequences. The 2D CNNs with pre-trained ImageNet weights successfully understand  spatiotemporal images. Against our expectations, spatial 2D CNNs performed better than spatiotemporal 3D CNNs on video datasets~\cite{JiPAMI2013,TranICCV2015}. Based on a previous report  ~\cite{HaraCVPR2018}, 3D CNNs require a large number of labeled videos to optimize the 3D kernels. In the same study, it was found that the Kinetics-400 dataset~\cite{KayarXiv2017} was able to successfully train 3D CNNs. Along these lines, the most recent trend is shifting to 3D CNNs, such as Inflated 3-Dimentional convolutional network (I3D)~\cite{CarreiraCVPR2017} and 3D Residual Network (3D ResNet)~\cite{HaraCVPR2018}. The 3D CNNs directly compute video volumes with spatiotemporal $xyt$ kernels. 

\cHiro{Although the spatiotemporal 3D CNNs heavily rely on the architecture modification, the recently released larger-scale datasets, e.g., Kinetics-700~\cite{Carreiraarxiv2019} and Moments in Time (MiT), allow us to have a great potential to further improve 3D CNNs. Here, we must consider how to efficiently use large-scale video datasets.}

Therefore, in the present paper, we disclose practical knowledge through an experimental study for video recognition. Here, we describe how to use and increase  a large-scale video dataset. Basically, we conducted pre-training on Kinetics-700~\cite{Carreiraarxiv2019}, MiT~\cite{Abu-El-Haijaarxiv2016}, and STAIR Action (STAIR)~\cite{Yoshikawaarxiv2018}, in addition to fine-tuning then UCF-101~\cite{Soomro2012}, HMDB-51~\cite{Kuehne2011}, and ActivityNet~\cite{HeilbronCVPR2015}.

We summarize our experiments and the knowledge obtained in the trials as follows.

\begin{itemize}
    \vspace{-8pt}\item At the beginning of our exploration study, to simply confirm the effects of pre-training, we conducted training and fine-tuning on Kinetics-700, MiT, STAIR, and the Mini-Holistic Video Understanding dataset (Mini-HVU). Although the MiT dataset contains 802k labeled videos in the training set, the Kinetics-700 (545k training videos) pre-trained model records better scores with 92.0\% for UCF-101, 66.0\% for HMDB-51, and 75.9\% for ActivityNet in top-1 video-level accuracy (see Section~\ref{sec:pretraining_datasets}).
    \vspace{-8pt}\item We attempt to clarify the relationship between the data amount and the video recognition performance. Following a comprehensive study of transfer learning on ImageNet~\cite{HuhNIPS2016WS}, we investigate the relationships between recognition accuracy and data amount using \{10, 30, 50, 70, 90\}\% of the full-dataset configuration, in addition to the training from scratch. As reported in the experiment, pre-training is better than training from scratch, even if only 10\% of the data of \#category and \#instance in Kinetics-700 and MiT are used. Moreover, \#category should be initially fixed, and \#instance is then increased on the video dataset (see Section~\ref{sec:dataamount}).
    \vspace{-8pt}\item In order to practically increase the data amount, we simply merge a couple of datasets, such as Kinetics-700 (700 categories in 545k training videos) + MiT (802k training videos in 339 categories), to contain 1.34M videos in 1,039 categories. The merged dataset with Kinetics-700 and MiT enhances the performance rate, +0.9 on UCF-101, +3.4 on HMDB-51, and +1.1 on ActivityNet, compared to the Kinetics-700 pre-trained model (see Section~\ref{sec:mergeddataset}).
    \vspace{-8pt}\item We investigate the effect of \#layer increase on Kinetics-400/700 and the merged dataset with Kinetics-700 and MiT by following a previous paper~\cite{HaraCVPR2018}. In the experiment, we also use (2+1)D CNNs~\cite{TranCVPR2018} as well as spatiotemporal 3D CNNs. Although the Kinetics-400 pre-trained 3D CNN decreases the transferred accuracy on ResNet-200, the Kinetics-700 pre-trained 3D CNN successfully performs training in the same configuration (see Figure~\ref{fig:depth_ucf}). The merged dataset pre-trained 3D CNN further strengthens the performance in video classification. In contrast, none of the pre-trained (2+1)D CNNs with ResNet-200 can be optimized in the fine-tuning phase (see Section~\ref{sec:deeperarch}).
\end{itemize}

More detailed results are shown in the experimental section. The reminder of the present paper is organized as follows. In Section 2, we introduce related research and the position of the present paper. The detailed experimental setting is described in Sections 3. The experimental results are presented and discussed in Section 4. Finally, we summarize the present paper in Section 5.



\section{Related work}

The present paper handles a topic in terms of spatiotemporal visual recognition for transfer learning in videos. Therefore, recent research is closely related to such as spatiotemporal models, large-scale video datasets, and an exploration study in CNN-based transfer learning. We list representative papers for each topic, as follows.

\vspace{-12pt}\paragraph{Exploration study in transfer learning.} 

As the researchers discussed, transfer learning with an well-organized dataset (e.g., ImageNet~\cite{DengCVPR2009} and Instagram-3.5B~\cite{ig3.5b}) and a sophisticated CNN architecture has allowed us to successfully recognize various objects, including humans~\cite{DonahueICML2014,KornblithCVPR2019}. Moreover, an exhaustive exploration study has been carried out in the context of image recognition. These efforts are highly beneficial in order to disclose practical knowledge and fair comparison with several approaches.

To the best of our knowledge, there are few discussions on comprehensive evaluation in video recognition. On one hand, several studies have focused on transfer learning in image classification. For example, Huh~\textit{et al.} evaluated several aspects of ImageNet transfer in terms of the relationship between the number of categories/instances~\cite{HuhNIPS2016WS}, and Kornblith~\textit{et al.} assessed why the ImageNet pre-training is so strong~\cite{KornblithCVPR2019}. This type of knowledge has helped to highly accelerate promising research in CNN-based image classification. 

We believe that the consideration of video transfer learning allowed recent video recognition to be more reliable and knowledgeable. We would like to validate several relationships, such as those among accuracy and \#instance/\#category (refer to \cite{HuhNIPS2016WS}), pre-training, and fine-tuning (refer to \cite{KornblithCVPR2019}), and we validate the importance of a simply increased \#video for training more deeper 3D CNNs. Although Hara~\textit{et al.} reported that 3D CNNs saturate 152 layers with ResNet~\cite{HaraCVPR2018}, we try to successfully optimize a ResNet with more deeper layers on recently proposed large-scale video datasets. 

\vspace{-12pt}\paragraph{Spatiotemporal models.} Early in the field of video recognition, the tracking of spatiotemporal points has become an epoch-making idea through sparse- (e.g., STIP~\cite{LaptevICCV2003,LaptevIJCV2005}) and dense-point detection (e.g., Dense Trajectories~\cite{WangCVPR2011,WangICCV2013}). In the era of the deep neural network, there are primarily three different approaches to extract a video representation: 2D CNNs (e.g., Two-Stream ConvNets~\cite{SimonyanNIPS2014,FeichtenhoferCVPR2016}, Temporal Segment Networks (TSN)~\cite{WangECCV2016}), and 3D CNNs (e.g., C3D~\cite{TranICCV2015}, I3D~\cite{CarreiraCVPR2017}, 3D ResNets~\cite{HaraCVPR2018}) and (2+1)D CNN (e.g., P3D~\cite{QiuICCV2017}, R(2+1)D~\cite{TranCVPR2018}).

Recently, 3D and (2+1)D CNNs have been said to be the most promising methods for video recognition. We basically conducted the experiments in the present paper using 3D-ResNet~\cite{HaraCVPR2018}. However, we compare these two methods in the last part of the experiment. In order to simply validate the effects of video datasets for 3D CNNs, we do not assign a stream with optical flows. In addition, we do not pursue state-of-the-art video classification performance in the present paper.

\vspace{-12pt}\paragraph{Video datasets.} We have witnessed several types of video datasets in terms of various domains and number of video data. Earlier, a couple of video datasets were proposed on visual surveillance and movie analysis (e.g., KTH~\cite{LaptevICCV2003}, Weizmann Actions~\cite{BlankICCV2005}, and Hollywood2~\cite{MarszalekCVPR2009}). 
Second, the video datasets by video sharing services (e.g., YouTube and Flickr) have been proposed, such as UCF-101, HMDB-51, and ActivityNet~\cite{Soomro2012,Kuehne2011,HeilbronCVPR2015}. Although these datasets are initially used as a training and validation set with a hand-crafted feature and classifier, we currently use the datasets in order to conduct fine-tuning with trained CNN models. Recently, as of 2019, video datasets are being highly increasing on video sharing platforms (to download video content) and crowdsourcing platforms (to annotate the videos). Along these lines, Sports-1M~\cite{KarpathyCVPR2014} and YouTube-8M~\cite{Abu-El-Haijaarxiv2016} have been constructed at as large a scale as possible with an automatically labeled video collection. However, automatic labels, such as user-defined meta data, are not well-organized ground truth in video recognition. On behalf of these huge datasets, Kinetics~\cite{KayarXiv2017,Carreiraarxiv2019} and Moments in Time (MiT)~\cite{Abu-El-Haijaarxiv2016} have replaced a pre-trained model in the context of a video dataset. In the present paper, we explore the question: ``Do recently released large-scale pre-trained video datasets further improve spatiotemporal 3D CNNs?". We also use STAIR Actions (STAIR)~\cite{Yoshikawaarxiv2018} and Holistic Video Understanding (HVU)~\cite{Dibaarxiv2019}, which contains approximately 100K well-labeled videos in addition to the above-mentioned Kinetics-700 and MiT datasets. In most cases of the evaluation, we use representative datasets by including UCF-101~\cite{Soomro2012}, HMDB-51~\cite{Kuehne2011}, ActivityNet~\cite{HeilbronCVPR2015}, and Kinetics-700~\cite{Carreiraarxiv2019}.

\cHiro{Moreover, we did not choose weakly supervised datasets, such as YouTube-8M~\cite{Abu-El-Haijaarxiv2016} and Sports-1M~\cite{KarpathyCVPR2014}, because we do not expect to achieve a higher-level performance. Though Ghadiyaram~\textit{et al.} achieved higher accuracy with a large number of weak labels, the Instagram-65M dataset is not publicly available~\cite{GhadiyaramCVPR2019}.}


\section{Experimental settings}

\subsection{Overview}
\label{sec:summary}
In order to simply disclose how to effectively use/create video datasets in 3D CNNs, we perform an exploration study of 3D CNNs on video datasets. \cHiro{We mainly} use the spatiotemporal 3D-ResNet~\cite{HaraCVPR2018} as a representative 3D CNN and Kinetics-700, MiT, STAIR, and HVU~\cite{Dibaarxiv2019} as pre-trained datasets, which contain around 100k videos with well-organized human annotations. Moreover, we assign UCF-101, HMDB-51, and ActivityNet for evaluation datasets. Here, our strategy in the exploration study is as follows: (i) What kind of dataset is suitable for transfer learning in video classification? In Section~\ref{sec:pretraining_datasets}, we compare multiple datasets in line with transfer learning. (ii) We would like to validate the relationship between data amount and the performance rate. In Section~\ref{sec:dataamount}, we list \#category and \#instance in relation to their corresponding accuracies. (iii) Using an existing dataset, a simply merged dataset is conducted in order to increase the video data amount. In Section~\ref{sec:mergeddataset}, we concatenate three datasets as four different patterns in order to improve their performances. (iv) Based on a previous paper~\cite{HaraCVPR2018}, we verify the relationship between \#layer and the performance increase in terms of the use of the ResNet architecture. In Section~\ref{sec:deeperarch}, we use ResNet-\{18, 34, 50, 101, 152, 200\} on Kinetics-400/700 and the best merged dataset shown in (iii). \cHiro{In the experiment (iv),} (2+1)D CNN is implemented for compare with the 3D CNN.



\subsection{Architectures}
The basic architecture of the 3D CNN is based on the 3D ResNet proposed by Hara~\textit{et al.}~\cite{HaraCVPR2018}. The procedure of video transfer learning consists of pre-training with large-scale datasets containing approximately 100k videos and fine-tuning by relatively small video datasets. At the beginning of experiment, we confirm the performance with 3D-ResNet-50 (Sections~\ref{sec:pretraining_datasets} -- \ref{sec:mergeddataset}) and add layers as with 3D-ResNet-\{18, 34, 50, 101, 152, 200\} (Section~\ref{sec:deeperarch}). 
Moreover, we consider (2+1)D CNNs~\cite{TranCVPR2018}, which have a similar philosophy to 3D CNNs. The (2+1)D convolution separately processes the spatial and temporal volume at each stacked block. 


\begin{table*}[t]
\begin{center}
\caption{Dataset details.}
\scalebox{1}{
\begin{tabular}{l|cccccc} \hline
  Dataset & Objective & Annotation type & Collection & \#Category & \#Video/\#Annotation & \#Video/\#Annotation \\
   & & & & & (train) & (validation) \\
      \hline \hline
  UCF-101 & Fine-tuning & Single label & YouTube & 101 & 9,537 / 9,537 & 3,783 / 3,783 \\
  HMDB-51 & Fine-tuning & Single label & Movie & 51 & 3,570 / 3,570 & 1,530 / 1,530 \\
  ActivityNet & Fine-tuning & Single label & YouTube & 200 & 10,024 / 10,024 & 4,926 / 4,926 \\
  Kinetics-700 & Pre-training & Single label & YouTube & 700 & 545,317 / 545,317 & 35,000 / 35,000 \\
  MiT & Pre-training & Single label & YouTube & 339 & 802,264 / 802,264 & 33,900 / 33,900 \\
  STAIR & Pre-training & Single label & User-defined & 100 & 99,478 / 99,478 & 10,000 / 10,000 \\
  Mini-HVU~\cite{Dibaarxiv2019} & Pre-training & Multiple labels & YouTube & 2,550 & 129,627 / \cHiro{3,192,077} & \cHiro{10,056} / \cHiro{209,702} \\
  HVU~\cite{Dibaarxiv2019}$^{*}$ & Pre-training & Multiple labels & YouTube & 4,378 & 481,418 / 11,902,432 & $^{**}$ \\
\hline
\multicolumn{6}{l}{* The full HVU dataset is not publicly available in the submission.}\\
\multicolumn{6}{l}{** The value is not reported in the paper~\cite{Dibaarxiv2019}.}\\
\end{tabular}
}
\label{tab:videodataset}
\end{center}
\vspace{-15pt}
\end{table*}

\subsection{Datasets}
The datasets used in the present paper are mainly divided into pre-training and fine-tuning datasets. We first introduce pre-trained datasets (Kinetics-700, MiT, STAIR, and HVU) and then describe evaluation (fine-tuning) datasets (UCF-101, HMDB-51, and ActivityNet). The listed datasets and their characteristics are shown in Table~\ref{tab:videodataset}.

\vspace{-10pt}\paragraph{Pre-training Datasets.} In order to successfully optimize convolutional kernels in 3D CNNs, the architecture basically requires a large amount of data. The number of video data are said to be over 100K. Therefore, we assign large-scale and easily available video datasets, namely Kinetics-700, MiT, STAIR Action, and Mini-HVU\footnote{We use mini-set of the HVU dataset (Mini-HVU) because the full HVU dataset was not publicly available when the paper was submitted.}~\cite{Dibaarxiv2019}. Compared to single-label datasets, the (Mini-)HVU dataset consists of multiple labels per video, which are based on scene, object, action, event, attribute, and concept. We calculate loss values with cross entropy loss and softmax function following Diba~\textit{et al.}~\cite{Dibaarxiv2019}. Therefore, the loss functions for these datasets are different from those for a single-label dataset with only cross entropy loss, due to the types of annotation. Several video datasets have been collected on video sharing sites, e.g., on YouTube. On the other hand, STAIR Actions has collected user-captured, user-labeled, and user-submitted videos on the cloud. 


\vspace{-10pt}\paragraph{Fine-tuning Datasets.} As we mentioned above, we mainly use UCF-101, HMDB-51, and ActivityNet, which are frequently used evaluation datasets in video recognition. That is, the three datasets are easily compared with conventional approaches. Most of these videos have been collected on YouTube, except for HMDB-51, which is downloaded from various cinemas. Moreover, we validate a score using larger-scale datasets (e.g., Kinetics-700) in Section~\ref{sec:comparison}. 

\subsection{Implementation details}


\paragraph{Training.} Basically, we use 3D-ResNet~\cite{HaraCVPR2018} for video classification tasks. Therefore, we follow the  configuration of parameters and training strategy. In addition, an input image sequence consists of 112 [pixel] $\times$ 112 [pixel] $\times$ 3 [channel] $\times$ 16 [frame] by cropping an input video. The 16-frame video clip is randomly cropped from a time position in the video. If the video sequence is shorter than 16 frames, the video clip is adjusted by iterating the video frames. In order to augment the training, we apply $\times$10 augmented images by adopting four-corner/center cropping and their horizontal flipping. We also consider the scale of video clips by multiplying \{1, $\frac{1}{2^{1/4}}$, $\frac{1}{\sqrt{2}}$, $\frac{1}{2^{3/4}}$, and $\frac{1}{2}$\}. Moreover, 10-crop augmentation and multi-scale sizing are randomly selected in mini-batch training, where we refer to the settings of Wang~\textit{et al.}~\cite{Wangarxiv2015_tscnn}.

In the training phase, we use stochastic gradient descent (SGD) and cross-entropy loss as an optimizer and a loss function, respectively. When we train a multi-label dataset on Mini-HVU, we assign a combined loss function with cross entropy and softmax loss by referring to Diba~\textit{et al.}~\cite{Dibaarxiv2019}. The weight decay and momentum are set as 0.001 and 0.9, respectively. The learning rate starts from 0.003 and is then updated if the validation loss is saturated for 10 epochs in a row. 

\begin{figure*}[t]
  \centering
  \subfigure[Fine-tuning on UCF-101]{\includegraphics[width=0.33\linewidth]{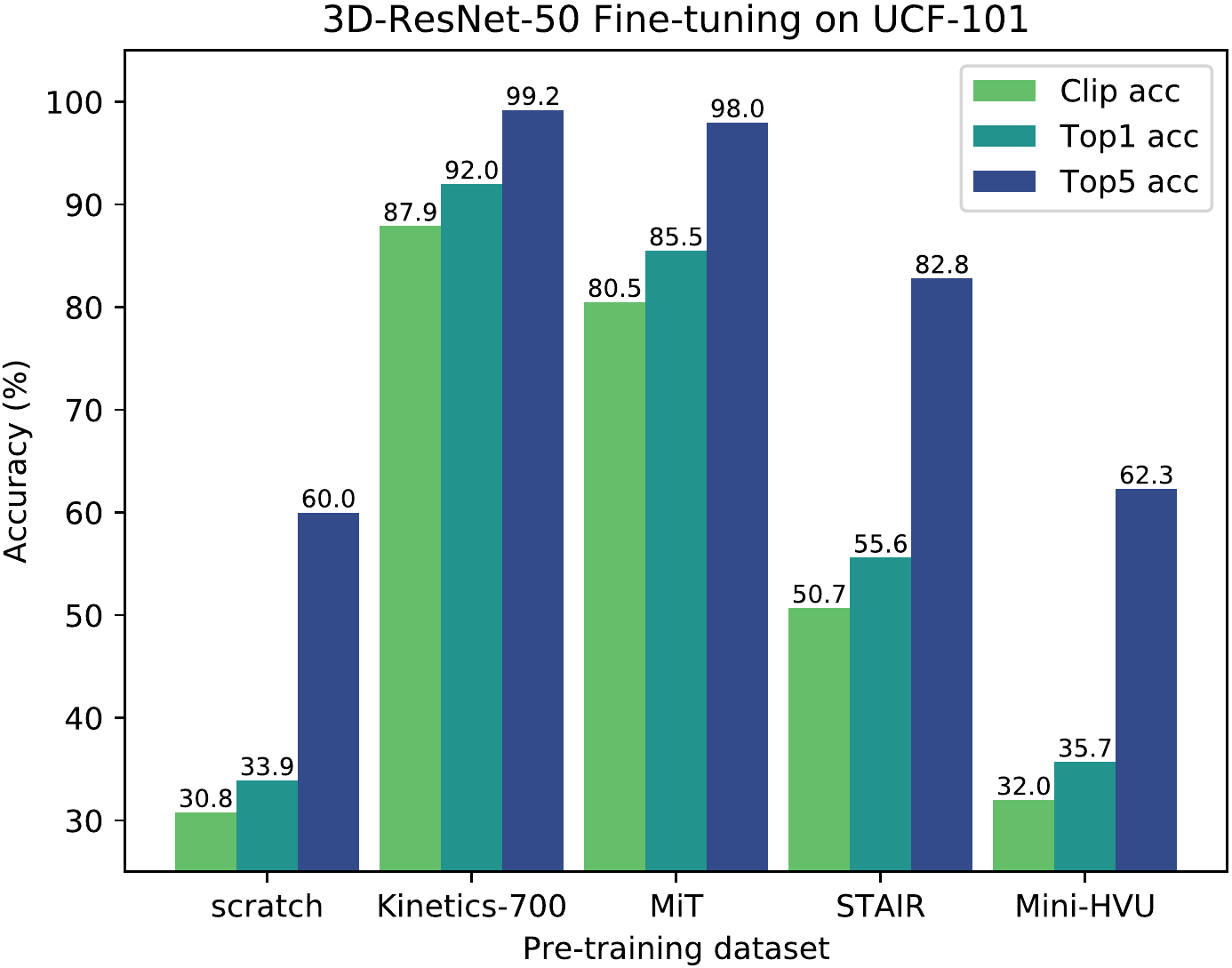}}
  \subfigure[Fine-tuning on HMDB-51]{\includegraphics[width=0.33\linewidth]{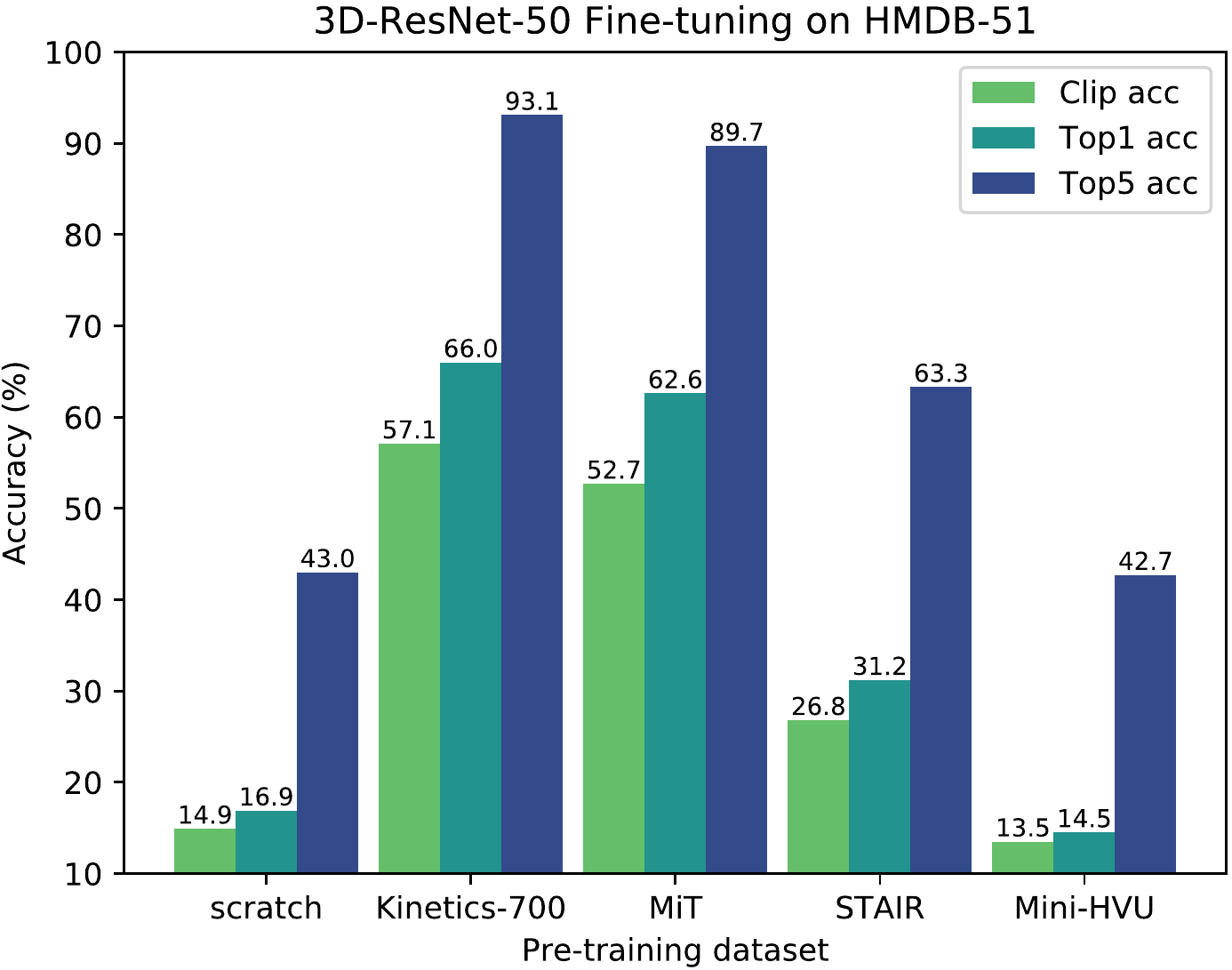}}
  \subfigure[Fine-tuning on ActivityNet]{\includegraphics[width=0.33\linewidth]{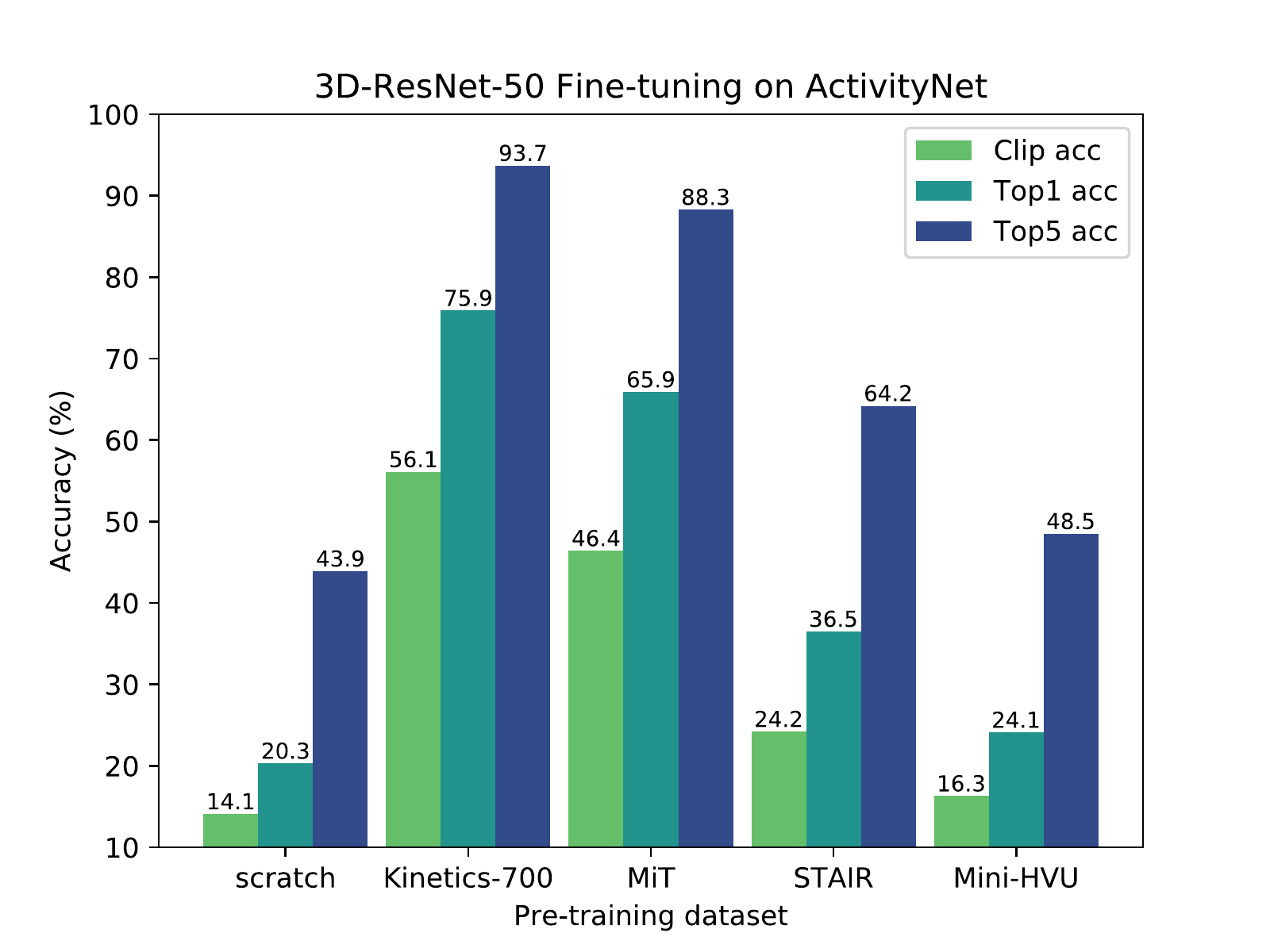}}
\vspace{-5pt}\caption{Transfer learning on video recognition datasets. The horizontal axes indicate the differences on a pre-training dataset.}
\label{fig:r50_ft}
\end{figure*}

\vspace{-10pt}\paragraph{Validation.} We calculate three types of video-related accuracy: the video clip, top-1, and top-5 video-level accuracy. The video clip accuracy is extracted from corrected prediction in all video clips, which includes 16 frames from a video. The top-1 video-level accuracy is totally evaluated by accumulating the probability of video clips in the video. In the same way, the top-5 accuracy is judged by ranked prediction. If there is a correct category in the top-5 prediction, the prediction is counted as the correct answer. Moreover, we use the non-overlapped sliding window approach to output probability with 3D-ResNet and perform accumulation in video order. We do not conduct a prediction-time data augmentation.

\section{Results and consideration}


\subsection{Pre-training with representative datasets}
\label{sec:pretraining_datasets}

In order to simply confirm the effects of pre-training in a single dataset, we conduct a transfer learning on pairs of \{Scratch, Kinetics-700, MiT, STAIR, Mini-HVU\}, and \{UCF-101, HMDB-51, ActivityNet\}.
Figure~\ref{fig:r50_ft} indicates the effects of training from scratch and pre-training with \{Kinetics-700, MiT, STAIR, Mini-HVU\} by comparing the results to scratch from random parameters. The scores for video clip and top-1/5 video-level accuracies are listed in each subfigure. As shown in the figure, the Kinetics-700 pre-trained model achieves the best performance rates in all transferred tasks. We confirm that the 3D-ResNet-50 records 92.0 on UCF-101, 66.0 on HMDB-51, and 75.9 on ActivityNet for top-1 video-level accuracy. The difference between Kinetics-700 and the second-best dataset, MiT, is 6.5 (92.0 - 85.5) on UCF-101, 3.4 (66.0 - 62.6) on HMDB-51, and 10.0 (75.9 - 65.9) on ActivityNet. This tendency is also observed in video clip and top-5 video-level accuracies. Although the \#instance in MiT (802k) is larger than that in Kinetics-700 (\cHiro{545K}), the Kinetics-700 pre-trainined model achieved better rates in video classification. The STAIR and Mini-HVU pre-trained 3D-ResNet-50 is better than training from scratch. However, the datasets contain fewer videos in pre-training. 
%
We receive the benefit of accuracy increase from pre-trained datasets. Hereafter, we assign top-3 pre-trained datasets, namely, Kinetics-700, MiT, and STAIR, by considering the huge computational time required and that the full-HVU dataset is not publicly available.

\begin{figure*}[t]
  \centering
  \subfigure[\#Category and video acc on UCF-101]{\includegraphics[width=0.32\linewidth]{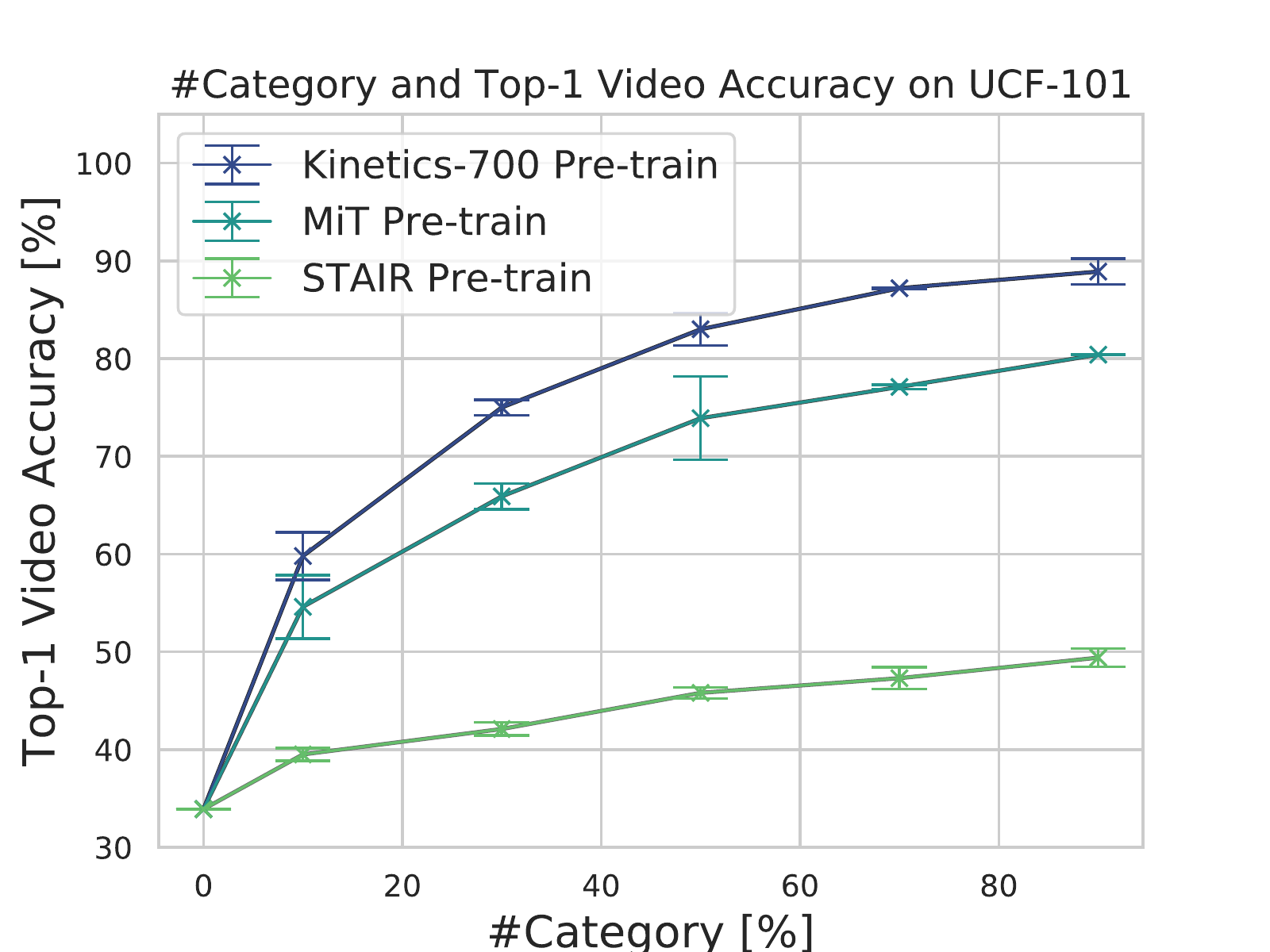}
  \label{fig:r50_cat_ucf}}
    \subfigure[\#Category and video acc on HMDB-51]{\includegraphics[width=0.32\linewidth]{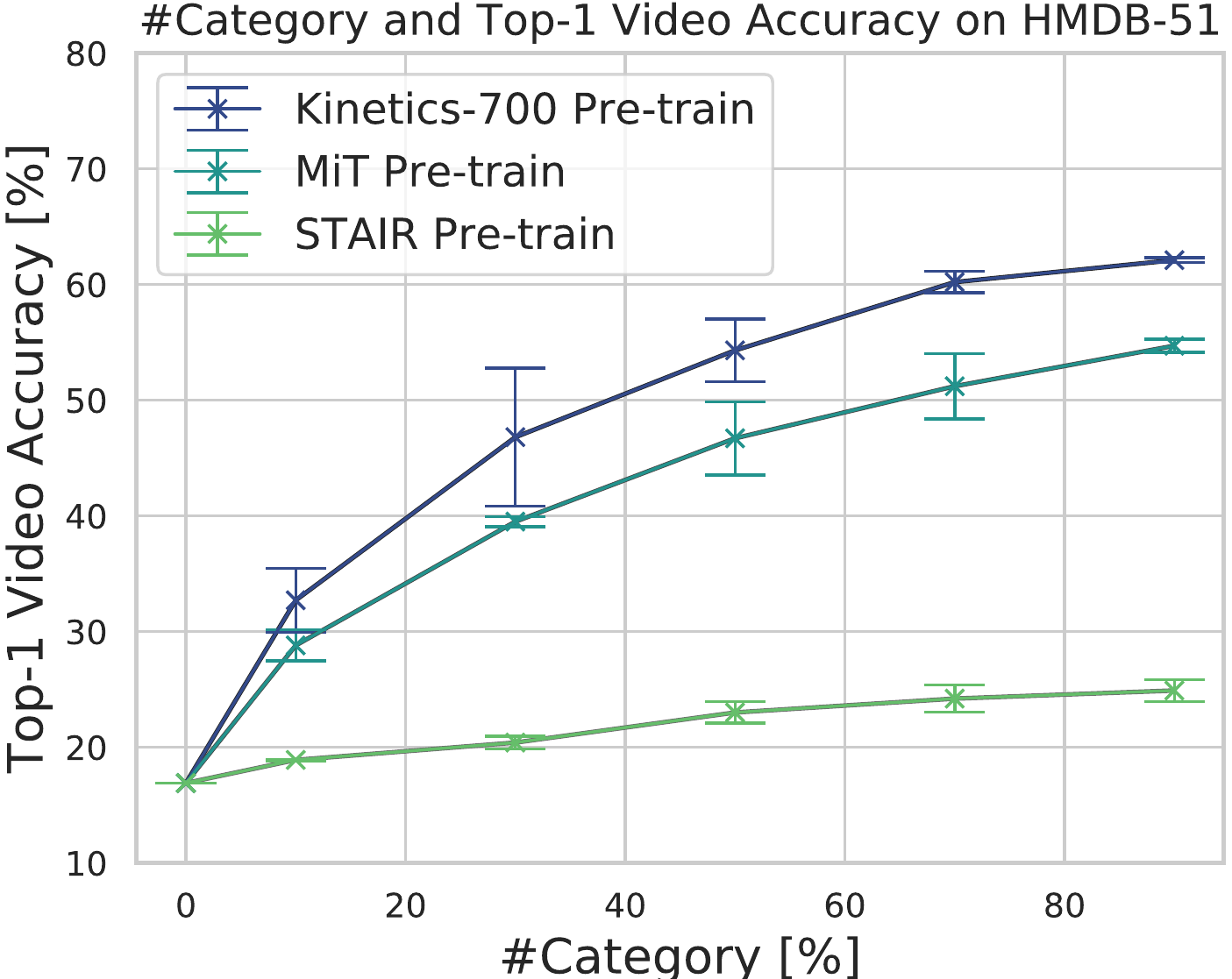}
  \label{fig:r50_cat_hmdb}}
  \subfigure[\#Category and video acc on ActivityNet]{\includegraphics[width=0.32\linewidth]{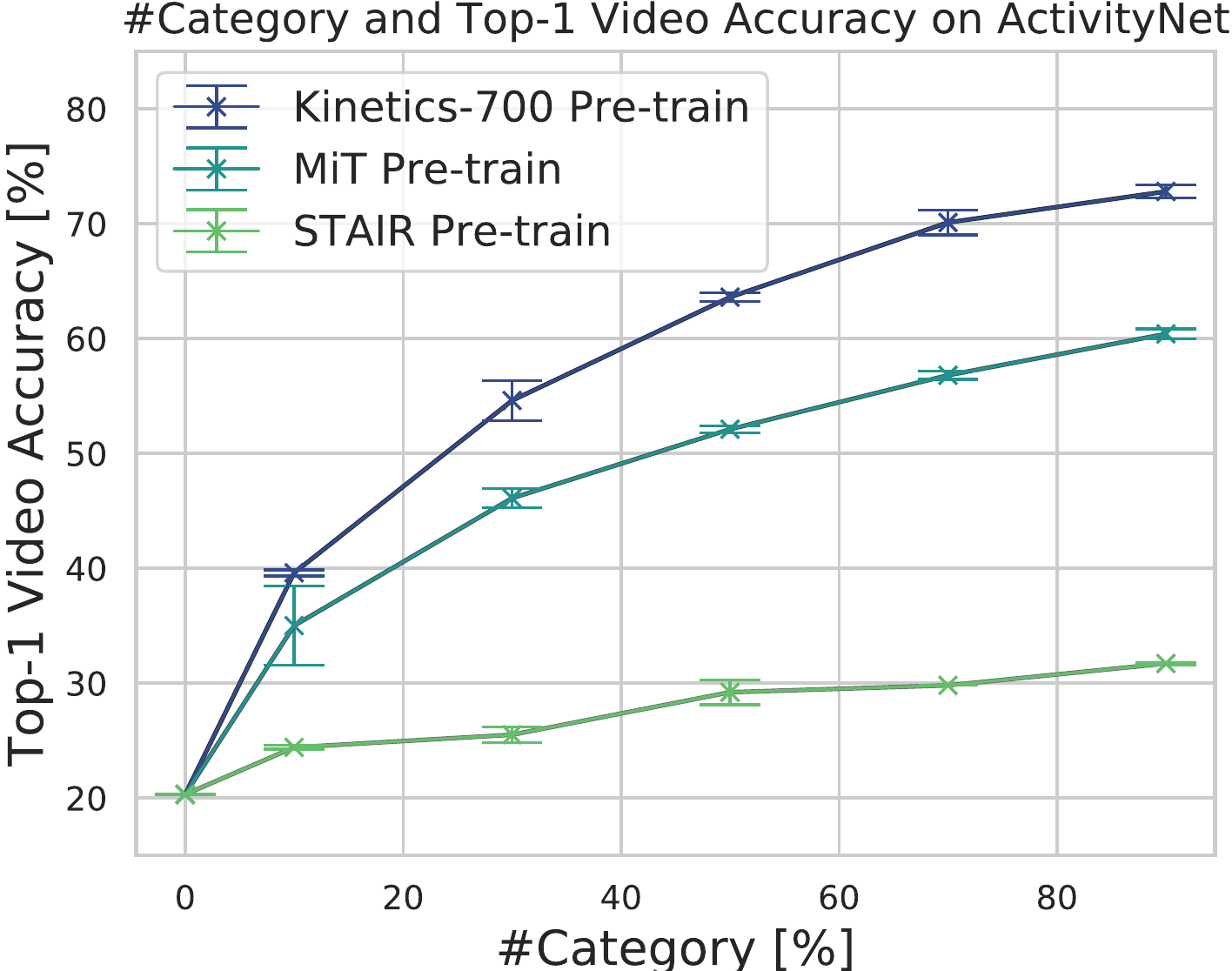}
  \label{fig:r50_cat_activitynet}}
  \subfigure[\#Instance and video acc on UCF-101]{\includegraphics[width=0.32\linewidth]{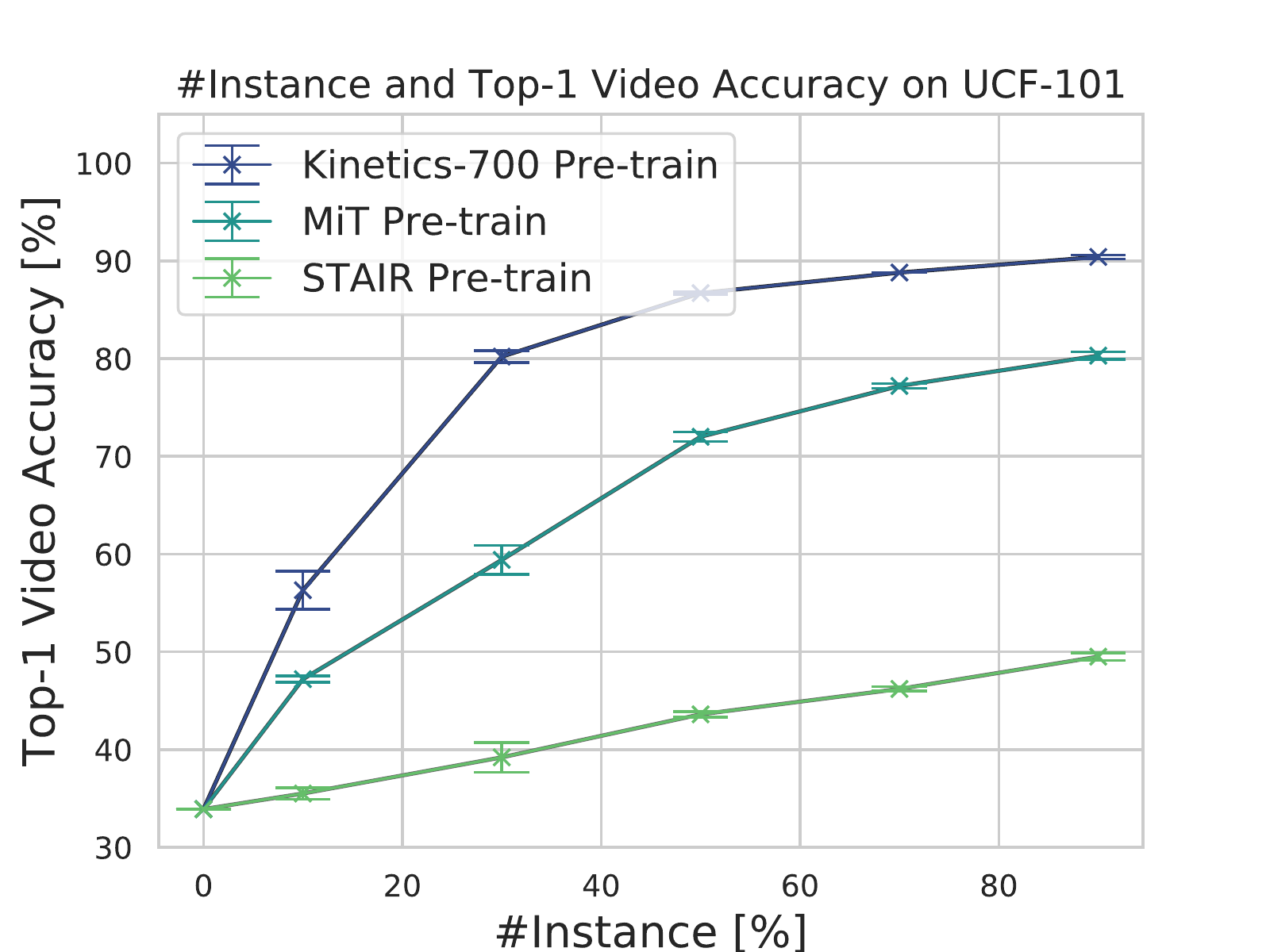}
  \label{fig:r50_ins_ucf}}
    \subfigure[\#Instance and video acc on HMDB-51]{\includegraphics[width=0.32\linewidth]{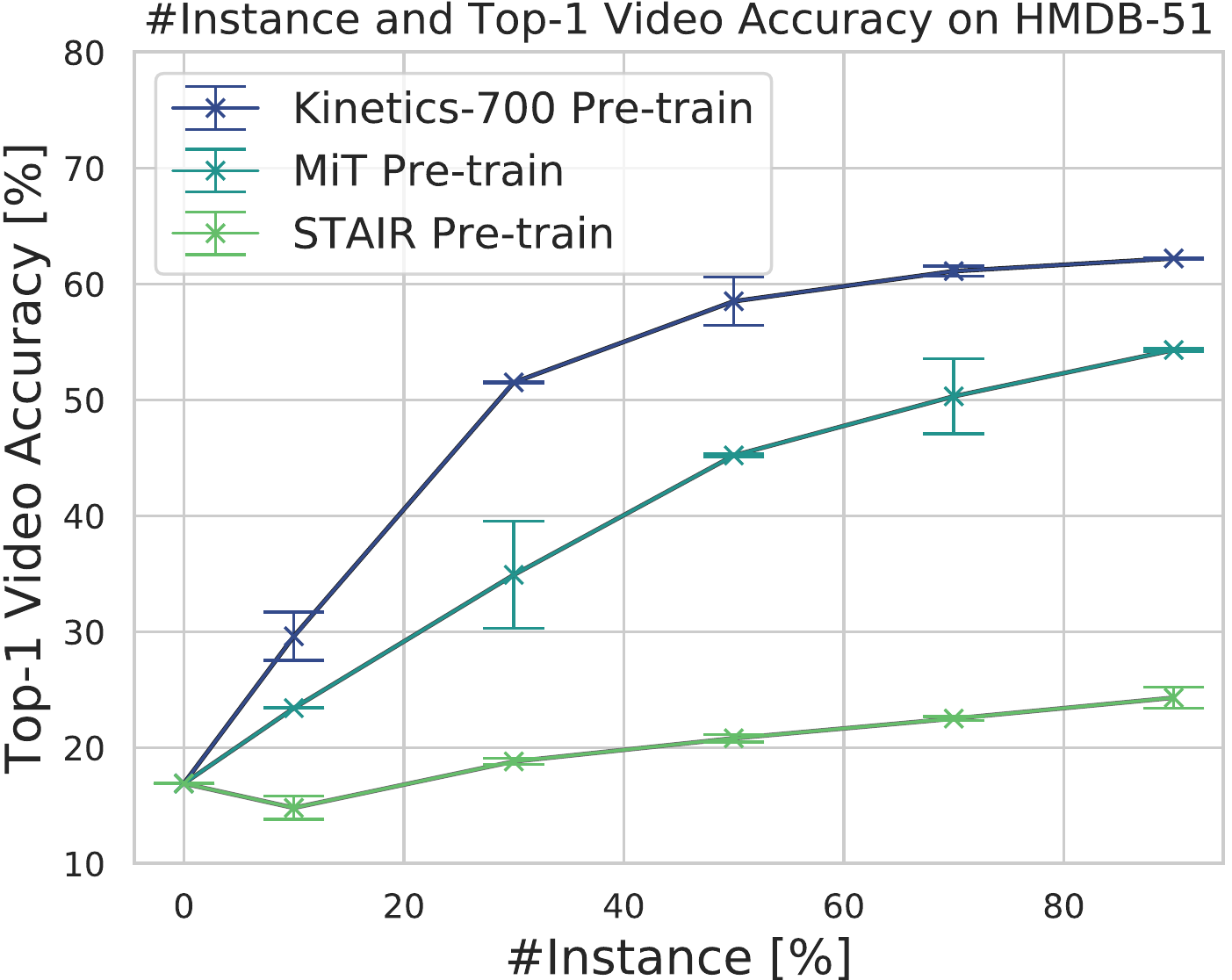}
  \label{fig:r50_ins_hmdb}}
  \subfigure[\#Instance and video acc on ActivityNet]{\includegraphics[width=0.32\linewidth]{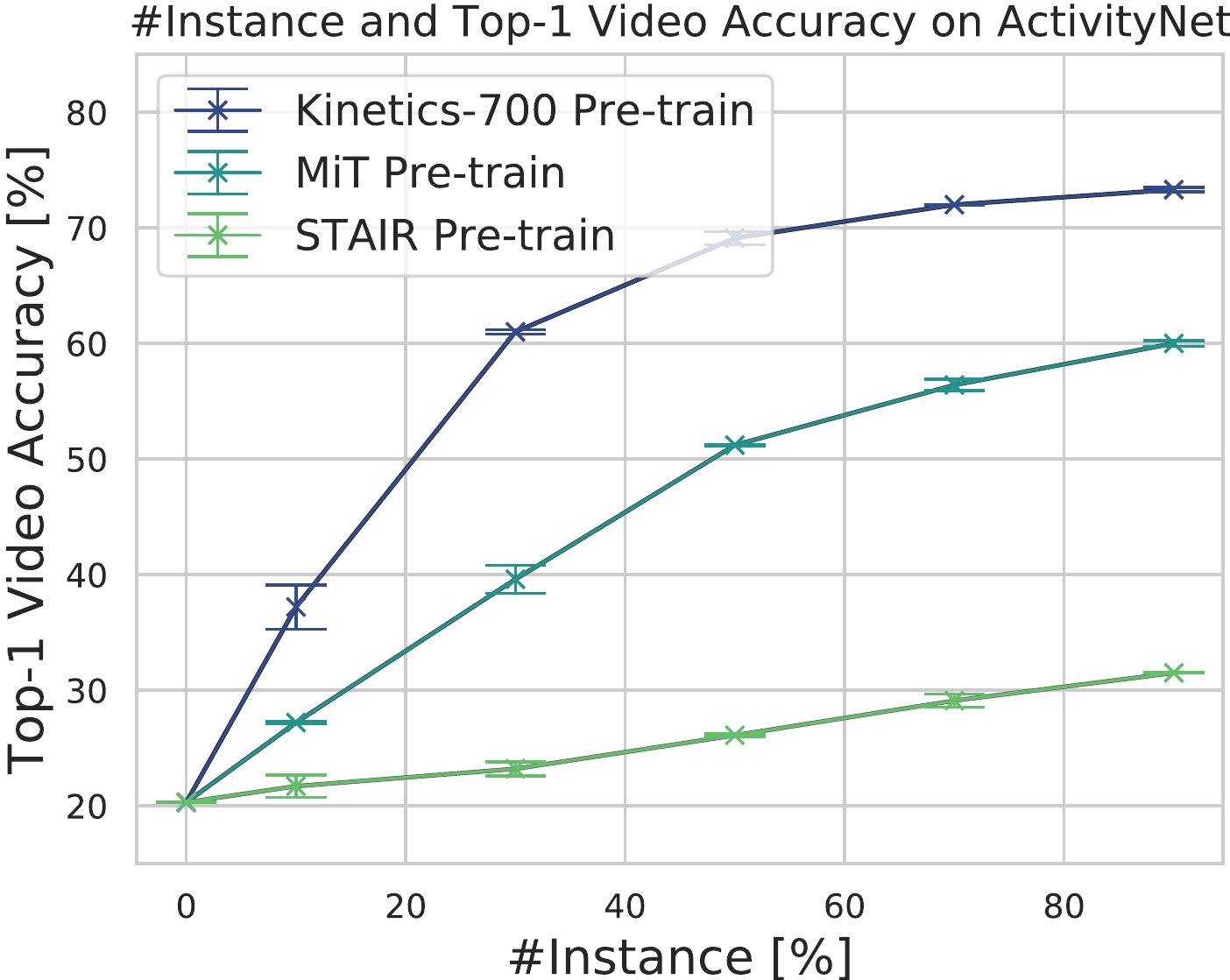}
  \label{fig:r50_ins_activitynet}}
\vspace{-5pt}\caption{Relationship between \#category/\#instance and top-1 video-level accuracy.}
\label{fig:cat_ins}
\end{figure*}

\begin{table*}[t]
\begin{center}
\caption{Merged datasets for pre-training. The uppercase characters indicate each dataset: K; Kinetics-700, M; MiT, and S; STAIR. For example, {\bf K+M} indicates  a merged dataset that combines {\bf K}inetics-700 and {\bf M}iT. We simply combine the \#video and \#category in pre-training. In this case, K+M contains 1,039 categories in 1.65M videos. The three different scores are indicated as the clip, top-1, and top-5 video-level accuracy.}
\scalebox{1}{
  \begin{tabular}{l|ccc} \hline
    Pre-train $\downarrow$ / Fine-tune $\rightarrow$ & UCF-101 & HMDB-51 & ActivityNet \\
     & Clip / Top-1 / Top-5 acc. & Clip / Top-1 / Top-5 acc. & Clip / Top-1 / Top-5 acc. \\
    \hline \hline
    K+M+S & 88.3 / 92.3 / {\bf 99.5} & 58.8 / 67.8 / 93.0 & 56.6 / 75.8 / 93.3 \\
    K+M & {\bf 89.1} / {\bf 92.9} / 99.4 & {\bf 60.4} / {\bf 69.4} / {\bf 94.0} & {\bf 57.4} / {\bf 77.0} / {\bf 93.9} \\
    K+S & 87.1 / 91.0 / 98.9 & 57.0 / 64.9 / 91.3 & 56.0 / 74.9 / 92.5 \\
    M+S & 76.4 / 81.3 / 96.1 & 48.9 / 56.4 / 84.7 & 43.5 / 62.8 / 86.3 \\
    Kinetics-700 (baseline) & 87.9 / 92.0 / 99.2 & 57.1 / 66.0 / 93.1 & 56.1 / 75.9 / 93.7 \\
    MiT (baseline) & 80.5 / 85.5 / 98.0 & 52.7 / 62.6 / 89.7 & 46.4 / 65.9 / 88.3 \\
    STAIR (baseline) & 50.7 / 55.6 / 82.8 & 26.8 / 31.2 / 63.3 & 24.2 / 36.5 / 64.2 \\
\hline
  \end{tabular}
 }
 \label{tab:mergeddataset}
\end{center}
\vspace{-20pt}
\end{table*}

\begin{figure*}[t]
  \centering
  \subfigure[Fine-tuning on HMDB-51]{\includegraphics[width=0.32\linewidth]{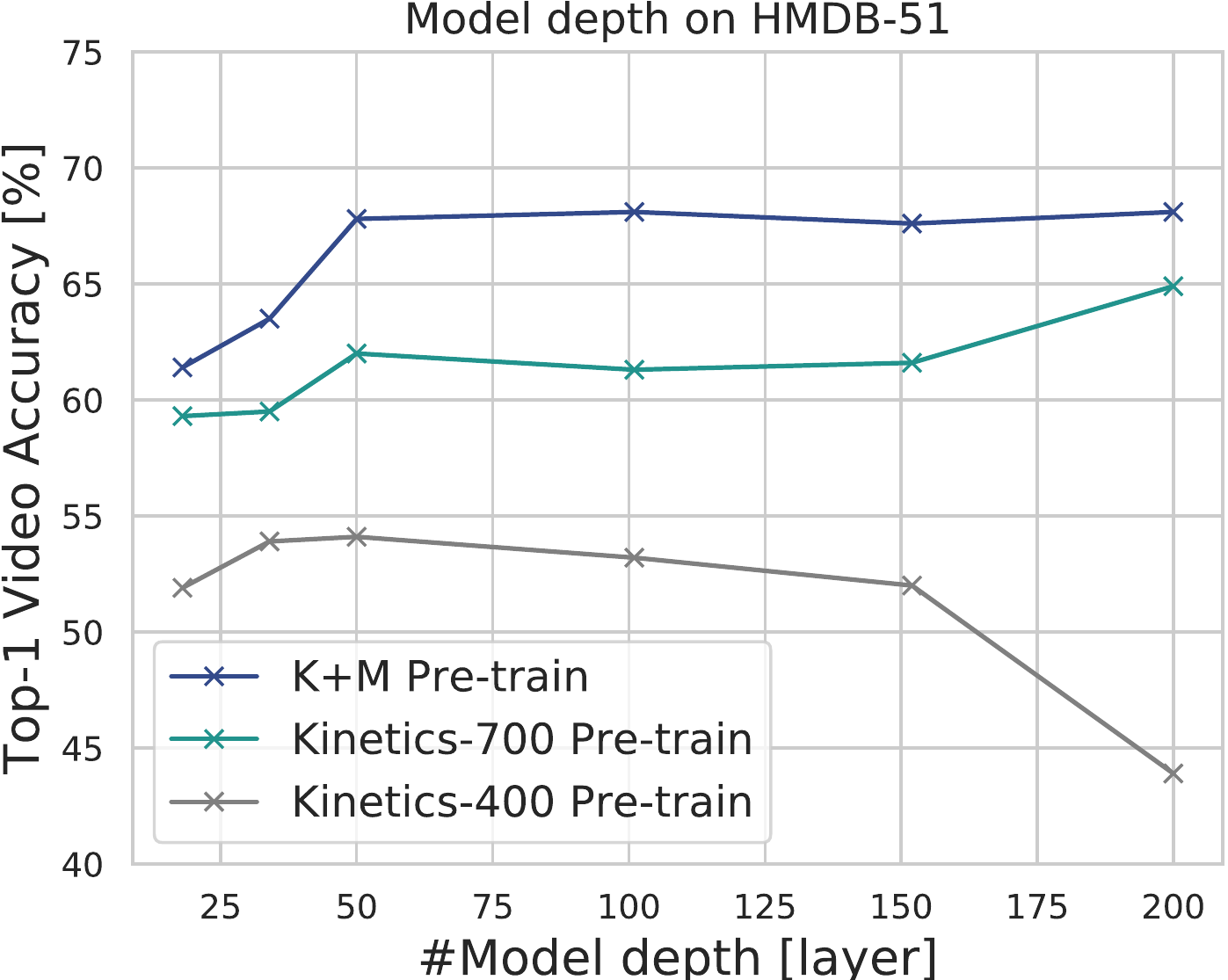}
  \label{fig:depth_hmdb}}
  \subfigure[Fine-tuning on ActivityNet]{\includegraphics[width=0.32\linewidth]{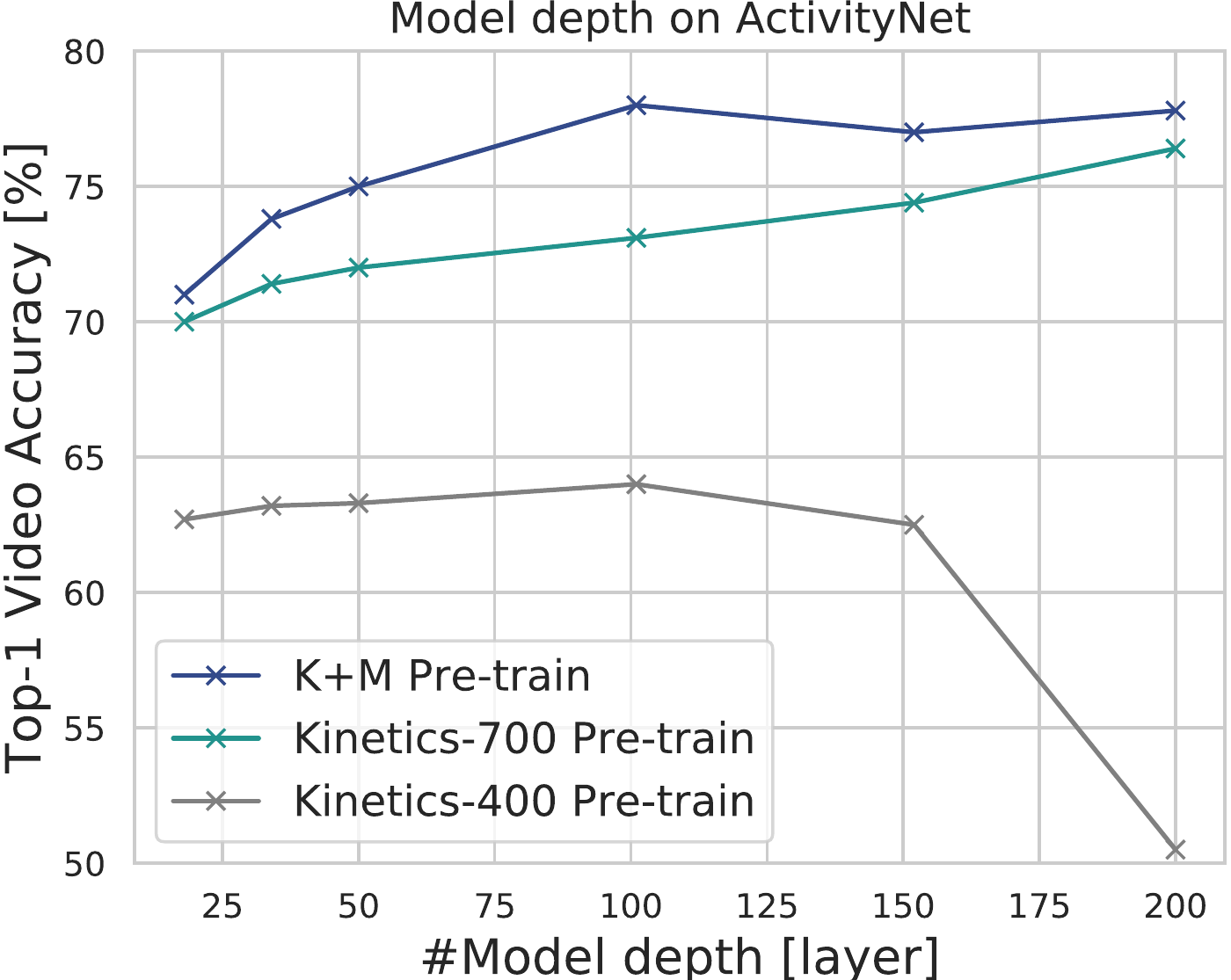}
  \label{fig:depth_activitynet}}
  \subfigure[Fine-tuning on Kinetics-700]{\includegraphics[width=0.32\linewidth]{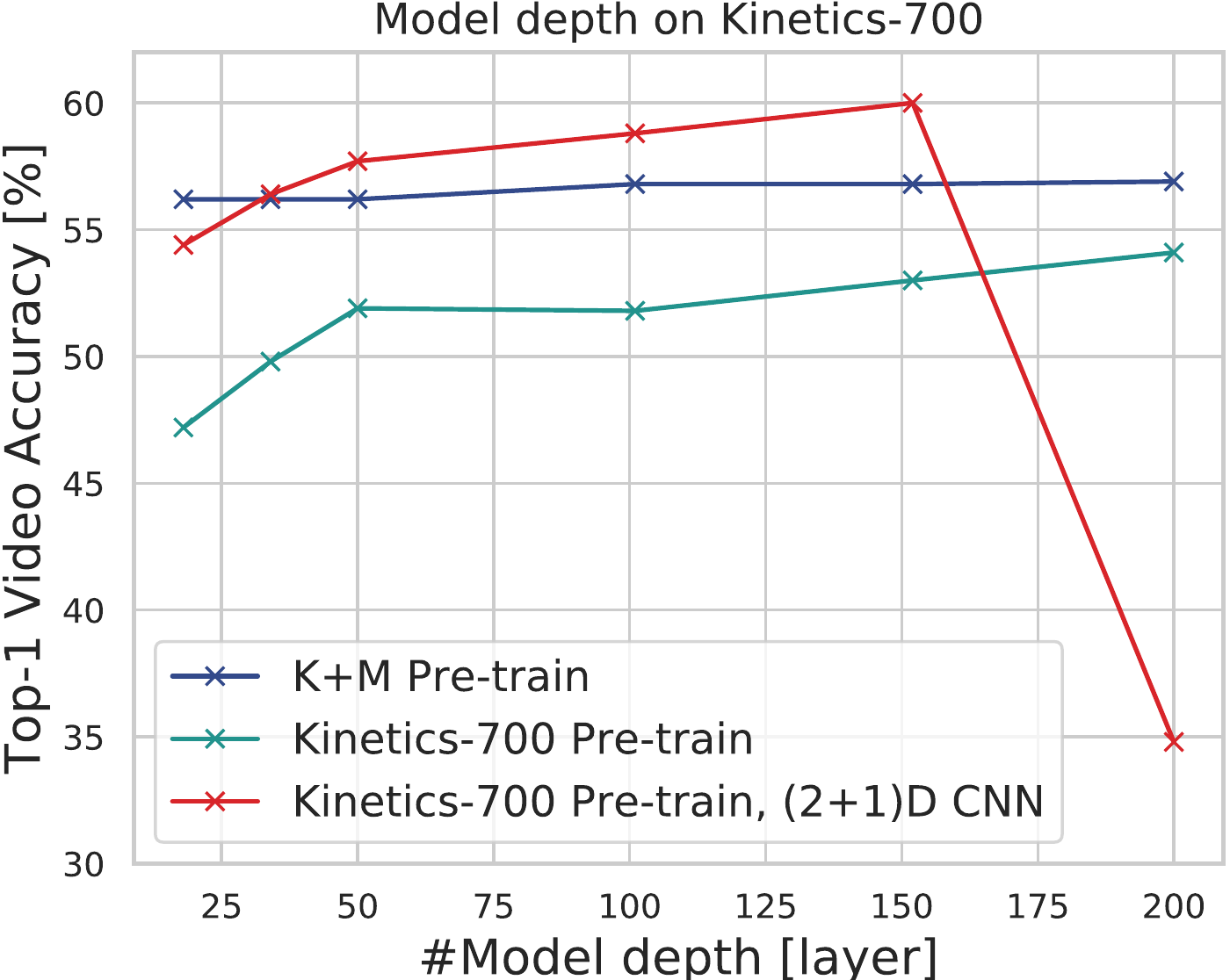}
  \label{fig:depth_kinetics}}
\vspace{-5pt}\caption{Effects of depth increase in 3D-ResNet-\{18, 34, 50, 101, 152, 200\} on all datasets and (2+1)D-ResNet-\{18, 34, 50, 101, 152, 200\} on Kinetics-700.}
\label{fig:depth}
\vspace{-10pt}
\end{figure*}

\subsection{Data amount (\#category/\#instance)}
\label{sec:dataamount}

In order to clarify the data amount and classification rates in the video dataset, we train and fine-tune various configurations in terms of the dataset.
Figure~\ref{fig:cat_ins} illustrates the relationships between \#category/\#instance and top-1 video-level accuracy. In comparing the two graphs (Figures~\ref{fig:cat_ins}(a)\& \ref{fig:cat_ins}(d), \ref{fig:cat_ins}(b)\& \ref{fig:cat_ins}(e), and \ref{fig:cat_ins}(c)\& \ref{fig:cat_ins}(f)), the increase in  \#category tends to improve the video accuracy. In other words, we should add category at the beginning. We intended to fix \#category and increase video instances (Figures~\ref{fig:cat_ins}(d), \ref{fig:cat_ins}(e), and \ref{fig:cat_ins}(f)) is faster training (Figures~\ref{fig:cat_ins}(a), \ref{fig:cat_ins}(b), and \ref{fig:cat_ins}(c)) on UCF-101, HMDB-51, and ActivityNet, respectively.

From another perspective, a larger dataset is not always beneficial in video transfer learning. For example, the MiT dataset is larger than Kinetics-700, yet the accuracy increase rate is as steep. The comparison of the training set size is 802k on MiT vs. 545k on Kinetics-700. Kinetics-700 dataset serves as a sophisticated video dataset. 

Here, we consider one reason why the STAIR pre-trained model provides lower performance rates due to containing a relatively small amount of data. The dataset contains 100K videos. As reported in another paper by Huh~\textit{et al.}~\cite{HuhNIPS2016WS}, a small amount of data yields a small benefit from pre-training. 


\subsection{Merged dataset}
\label{sec:mergeddataset}

We would like to simply and practically increase the video recognition accuracy with public datasets. Using two or three datasets, we try to organize more larger datasets, such as {\bf K}inetics-700 + {\bf M}iT (K+M). Here, we simply concatenate two different datasets from 650K videos/700 categories and 1M videos/339 categories into 1.65M videos/1,039 categories.

We list three baselines (the scores are also shown in Figure~\ref{fig:r50_ft}) and four simply merged datasets in Table~\ref{tab:mergeddataset}. We considered three datasets as well as the experiment on the data amount in Section~\ref{sec:dataamount}. As reported in Table~\ref{tab:mergeddataset}, K+M pre-training achieved the best scores, as compared to pre-training with concatenated K+M+S dataset. By comparing the baseline Kinetics-700 pre-training on top-1 video-level accuracy, the gap is +0.9, +3.4, and +1.1 for UCF-101, HMDB-51, and ActivityNet, respectively. On the other hand, the accuracy of the K+M+S pre-trained model decreased slightly compared to that of the K+M pre-trained model. 
The result is different from the tendency for increased data to provide increased accuracy. One reason for this is that STAIR is a collection of user-defined videos on the cloud.  The domain is different from fine-tuning video datasets that contain YouTube-related UCF-101/Activitynet and movie-based HMDB-51. 

Based on the results of the experiment, we must consider the domain of the fine-tuning task. Merged datasets do not always work well in video classification.

\subsection{Increase in the number of model layers}
\label{sec:deeperarch}

We validate the relationship between the number of layers in ResNet and the video recognition accuracy on UCF-101, HMDB-51, ActivityNet, and Kinetics-700. As mentioned in 3D-ResNet~\cite{HaraCVPR2018}, Kinetics-400 pre-trained ResNet-152 is saturated for the video classification task. Moreover, we verify the fine-tuning accuracy on UCF-101, HMDB-51, and ActivityNet with the Kinetics-400/700 and K+M pre-trained models. 

Figures~\ref{fig:depth_ucf} and~\ref{fig:depth} depict the relationships on UCF-101 (Figure~\ref{fig:depth_ucf}), HMDB-51 (Figure~\ref{fig:depth_hmdb}), ActivityNet (Figure~\ref{fig:depth_activitynet}), and Kinetics-700 (Figure~\ref{fig:depth_kinetics}). The results for the fine-tuning datasets (UCF-101, HMDB-51, and ActivityNet) reveal that the accuracy of Kinetics-400 pre-trained 3D-ResNet-200 decreased, even though the accuracies of the Kinetics-700 and K+M pre-trained models improved slightly. The combination of Kinetics-400 and 3D-ResNet-200 cannot be optimized in the fine-tuning tasks. 

The accuracy gap between the Kinetics-400 and Kinetics-700 pre-trained models is approximately +3 -- 7\%. A significant improvement comes from video collection (from 300K to 650K videos in total), human annotation (including cross-check), and a carefully defined category. We confirm that Kinetics-700 pre-trained ResNet-200 overcomes the shrinkage on UCF-101, HMDB-51, and ActivityNet. The fine-tuning accuracy is improved in the deepest 200-layer ResNet. 

Moreover, the K+M pre-trained model perform better than the Kinetics-700 pre-trained model in Figure~\ref{fig:depth}. The simply merged datasets in terms of category and instance improve the fine-tuning accuracy. The simple yet practical approach an improvement of at most +3.1 on UCF-101, +6.0 on HMDB-51, and +4.9 on ActivitiNet compared to Kinetics-700 datasets. 

\begin{table*}[t]
\begin{center}
\caption{Methods, pre-training datasets, and fine-tuning datasets.}
\scalebox{1}{
  \begin{tabular}{l|c|cccc} \hline
    Method-\#layer & Pre-training (\#training-video) & UCF-101 & HMDB-51 & ActivityNet & Kinetics-700 \\
    \hline \hline
    R3D-34~\cite{Tranarxiv2017} & Sports-1M (793K) & 85.8 & 54.9 & -- & -- \\
    R3D-18~\cite{Dibaarxiv2019} & HVU (481k) & 90.4 & 65.1 & -- & -- \\
    R3D-18 & Kinetics-700 (545k) & 87.8 & 59.3 & 70.0 & 47.2\\
    R3D-34 & Kinetics-700 (545k) & 88.8 & 59.5 & 71.4 & 49.8\\
    R3D-50 & Kinetics-700 (545k) & 92.0 & 66.0 & 75.9 & 54.7\\
    R3D-200 & Kinetics-700 (545k) & 92.0 & 66.0 & 75.9 & 54.1 \\
    R3D-50 & K+M (1.34M) & 92.9 & 69.4 & 77.0 & 56.8 \\
    R3D-200 & K+M (1.34M) & 92.0 & 68.1 & 77.8 & 56.9 \\
    R(2+1)D-50 & Kinetics-700 (545k) & 93.4 & 69.4 & 78.4 & 57.7 \\
    \cHiro{R(2+1)D-200} & \cHiro{Kinetics-700 (545k)} & \cHiro{78.5} & \cHiro{50.5} & \cHiro{54.8} & \cHiro{34.8} \\
    \cHiro{R(2+1)D-50} & \cHiro{K+M (1.34M)} & \cHiro{91.2} & \cHiro{66.4} & \cHiro{74.0} & \cHiro{55.1} \\
    \cHiro{R(2+1)D-200} & \cHiro{K+M (1.34M)} & \cHiro{79.5} & \cHiro{52.9} & \cHiro{58.9} & \cHiro{40.6} \\\hline  \end{tabular}
 }
 \label{tab:comparisons}
\end{center}
\vspace{-20pt}
\end{table*}

\begin{figure}[t]
  \centering
  \subfigure[Training from scratch.]{\includegraphics[width=0.950\linewidth]{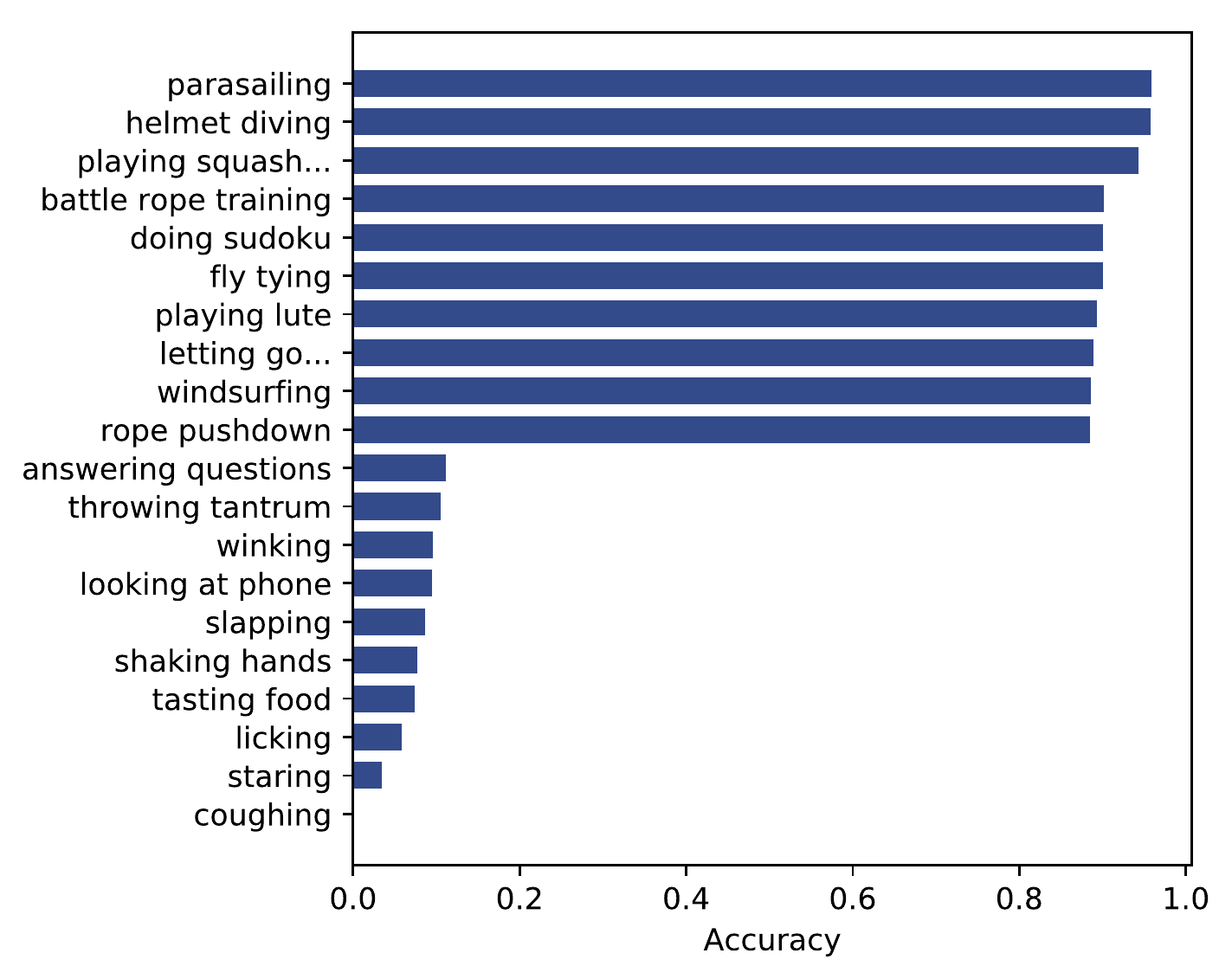}
  \label{fig:k2k_topbottom}}
  \subfigure[K+M pre-training.]{\includegraphics[width=0.950\linewidth]{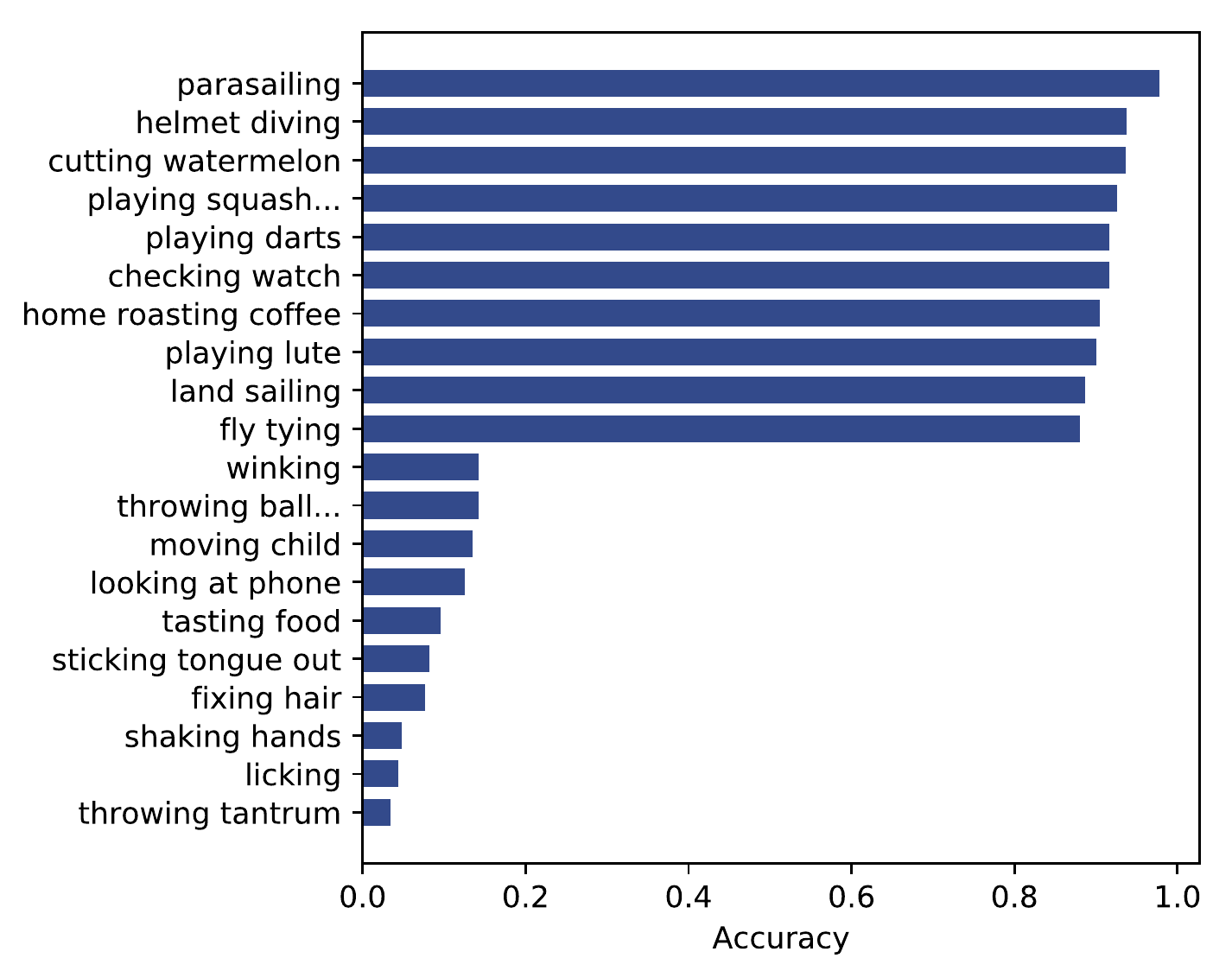}
  \label{fig:km2k_topbottom}}
\vspace{-0pt}\caption{\cHiro{Top/bottom-20 categories sorted by performance rates on Kinetics-700.}}
\label{fig:topbottom20}
\vspace{-20pt}
\end{figure}

\subsection{Comparison}
\label{sec:comparison}

In Table~\ref{tab:comparisons}, we list the ResNet-related 3D/(2+1)D CNNs for fair comparisons in terms of method, pre-training, backbone network, and fine-tuning datasets. We used reported scores in Sports-1M and HVU pre-trained 3D ResNets~\cite{Tranarxiv2017,Dibaarxiv2019}, as well as our exploration study.

For a pre-trained dataset, Kinetics-700 is better than Sports-1M (88.8 vs. 85.8 on UCF-101 and 59.5 vs. 54.9 on HMDB-51 based on ResNet-34) and yet is worse than HVU (87.8 vs. 90.4 on UCF-101 and 59.3 vs. 65.1 on HMDB-51 based on ResNet-34). Although Sports-1M contains a large number of labeled videos, automatic annotation was not enough to be better than Kinetics-700. Unlike our mini-HVU pre-training experiment in Figure~\ref{fig:r50_ft}, HVU pre-trained R3D-18 outperformed the same configuration with Kinetics-700. The significant data increase (129k$\rightarrow$481k) improves video classification accuracy in addition to the multi-label dataset. Moreover, a merged dataset with MiT enhances the Kinetics-700 pre-trained model (+0.9 on UCF-101, +3.4 on HMDB-51, +1.1 on ActivityNet, and +2.7 on Kinetics-700 based on R3D-50).

In terms of recognition architecture, R(2+1)D architectures provided a better recognition performance in the ResNet-50 backbone. However, R(2+1)D with ResNet-200 cannot be optimized in a given dataset. The performance rate of R(2+1)D decreased from 93.4 and 57.7 to 78.5 (-14.9) and 34.8 (-22.9) on UCF-101 and Kinetics-700, respectively.

\cHiro{Moreover, Figure~\ref{fig:topbottom20} lists top/bottom-20 categories sorted by top-1 video-level accuracy on Kinetics-700. Figure~\ref{fig:k2k_topbottom} and \ref{fig:km2k_topbottom} denote training from scratch and K+M pre-trained 3D-ResNet-50, respectively. We confirm that the bottom-20 categories are changed depending on the pre-training.}

\section{\cHiro{Discussion and} conclusion}

The present paper mainly revealed that 3D ResNets, including (2+1) ResNets, will further improve the video recognition accuracy by considering how to use video datasets. Through the comparison of representative video datasets, we showed that the Kinetics-700 and the merged Kinetics-700 + MiT pre-trained 3D-ResNet-200 are improved by fine-tuning tasks. Here, we summarize other knowledge through our experiments to transfer learning for video recognition.

\vspace{-15pt}\paragraph{The Kinetics pre-trained model is strong.} In Kinetics-700, the pre-trained model achieved better accuracy for single-dataset pre-training (see Figure~\ref{fig:r50_ft}). The improvement is +3.1 on UCF-101, +3.0 on HMDB-51, and +5.8 on ActivityNet compared to Kinetics-400 pre-training and +6.5, +3.4, and +5.4 compared to MiT pre-training (see Figure~\ref{fig:r50_ft}) based on 3D-ResNet-50. The results show that the amount of data is not all aspects,  namely, Kinetics-700 contains a relatively smaller number of videos (650k) compared to the one million videos of MiT. In order to normalize the dataset size between Kinetics-700 (545k training videos) and MiT (also, 802k), we compare an approximately equivalent data size with a 70\% \#instance amount on MiT (see Figure~\ref{fig:cat_ins}). The MiT configuration contains 574K training videos. In this case, the Kinetics-700 pre-trained model significantly improved MiT pre-training, i.e., 92.0 vs. 77.2 (+14.8) on UCF-101, 66.0 vs. 50.3 (+15.7) on HMDB-51, and 75.9 vs. 56.4 (+19.5) on ActivityNet. As mentioned in a previous study for the Kinetics dataset~\cite{KayarXiv2017}, ``three or more confirmations (out of five) were required before a clip was accepted" and ``classes were checked for overlap and de-noised". In other words, the Kinetics dataset was better in terms of human annotation because of a careful (re-)annotation of a large number of videos.


\vspace{-15pt}\paragraph{Data amount in pre-training.} What kind of dataset is required in terms of the video data amount? Through our exploration experiments, we assume that approximately 100K videos are not sufficient to pre-train a video dataset based on the fine-tuning results on Mini-HVU (129k) and STAIR (99k) pre-trained models. The STAIR-pre-trained 3D-ResNet-50 had a performance of 55.6/31.2 on UCF-101/HMDB-51. On the other hand, Kinetics-400 (which contains 240k training videos) pre-trained 3D-ResNet-50 provided a better accuracy, which was also reported in Hara~\textit{et al.}~\cite{HaraCVPR2018}. In their report, the Kinetics-400-pre-trained 3D-Resnet-50 had a performance of 89.3/61.0 on UCF-101/HMDB-51. Moreover, Kinetics-700 outperformed pre-training with other datasets. The Kinetics-700-pre-trained 3D-ResNet-50 achieved 92.0/66.0 on UCF-101/HMDB-51, which was the best score in single-dataset pre-training. Here, larger datasets, including MiT (802k) and Sports-1M (793K), cannot surpass the Kinetics-700 pre-trained model, as mentioned in the above discussion. Larger is not always better. The quality of human annotation is related to the pre-training in video classification.

According to the Figure~\ref{fig:cat_ins}, \#category is more important than \#instance in video recognition. We clarified the relationship between data amount and video recognition accuracy. Along these lines, we varied the number of categories and instances in pre-training. Here, we confirmed that a fixed \#category and a varied \#instance tends to increase the video recognition accuracy (see Figure~\ref{fig:cat_ins}). 


\vspace{-15pt}\paragraph{A merged dataset is one solution to increasing the amount of data available for training.} The results of the present study suggest that 3D CNNs can be improved by a merged dataset, for instance Kinetics-700 (700 categories in 650K videos) + MiT (339 categories in 1M videos) for 1,039 categories and 1.65M videos (K+M dataset). The merged K+M dataset helped to provide an improvement of +0.9 on UCF-101, +3.4 on HMDB-51, and +1.1 on ActivityNet from the Kinetics-700 pre-training. However, a merged dataset does not always provide better accuracy. The accuracy was decreased when we merged the dataset with STAIR in addition to the above-mentioned datasets. The gap between K+M{\bf+S} and K+M is -0.6, -1.6, and -1.2 on UCF-101, HMDB-51, and ActivityNet, respectively. We must consider the video domain in pre-training and fine-tuning. Unlike other pre-training datasets, STAIR has collected user-defined videos on the cloud. 

In the future, we intend to further improve 3D CNNs (2+1)D CNNs as datasets become larger. Moreover, we would like to find an easier and more practical approach to enhancing pre-trained 3D CNNs. 

\section*{Acknowledgment}

This work was supported by ABCI, AIST. We also want to thank Naoya Chiba and Ryosuke Araki for their helpful comments during research discussions. 


{\small
\bibliographystyle{ieee_fullname}
\bibliography{egbib}
}

\end{document}